\documentclass[sigconf, nonacm]{acmart}

\AtBeginDocument{%
  }

\usepackage{indentfirst}
\usepackage{color}
\usepackage{soul,xcolor}
\usepackage{amsmath} 
\usepackage{graphicx}
\usepackage{subcaption}
\usepackage{caption,subcaption}
\usepackage{multicol}
\usepackage[normalem]{ulem}
\usepackage{balance}
\usepackage[ruled, vlined, noend, linesnumbered]{algorithm2e}
\usepackage{lipsum}

\usepackage{wrapfig}

\newcommand{\algoshort}
{ACORD}
\newcommand{\algofull}{Adjustable Control Of RL Dynamics}
 \newcommand{\kname}{behavior oversight parameter}

\begin{document}

\author{Isaac Sheidlower}
\affiliation{%
  \institution{Tufts University}
  \city{Medford}
  \state{Massachusetts}
  \country{USA}}
\email{isaac.sheidlower@tufts.edu}

\author{Mavis Murdock}
\affiliation{%
  \institution{Tufts University}
  \city{Medford}
  \state{Massachusetts}
  \country{USA}}
\email{mavis.murdock@tufts.edu}

\author{Emma Bethel}
\affiliation{%
  \institution{Tufts University}
  \city{Medford}
  \state{Massachusetts}
  \country{USA}}
\email{emma.bethel@tufts.edu}

\author{Reuben M. Aronson}
\affiliation{%
  \institution{Tufts University}
  \city{Medford}
  \state{Massachusetts}
  \country{USA}}
\email{reuben.aronson@tufts.edu}

\author{Elaine Schaertl Short}
\affiliation{%
  \institution{Tufts University}
  \city{Medford}
  \state{Massachusetts}
  \country{USA}}
\email{elaine.short@tufts.edu}

\begin{abstract}
Reinforcement Learning (RL) is an effective method for robots to learn tasks. However, in typical RL, end-users have little to no control over \emph{how} the robot does the task after the robot has been deployed. To address this, we introduce the idea of \emph{online behavior modification}, a paradigm in which users have control over behavior features of a robot in real-time as it autonomously completes a task using an RL-trained policy. To show the value of this user-centered formulation for human-robot interaction, we present a behavior-diversity--based algorithm, Adjustable Control Of RL Dynamics
(ACORD), and demonstrate its applicability to online behavior modification in simulation and a user study. In the study (\textbf{n}=23), users adjust the style of paintings as a robot traces a shape autonomously. We compare ACORD to RL and Shared Autonomy (SA), and show ACORD affords user-preferred levels of control and expression, comparable to SA, but with the potential for autonomous execution and robustness of RL. The code for this paper is available at https://github.com/AABL-Lab/HRI2024\_ACORD

\end{abstract}

\keywords{Human-robot Interaction, User-centered Learning}
  
% [Towards] Expressive User Control of RL-Trained Robots
% Online Behavior Modification in RL: Enabling User Control of Robot Behavior Parameters
% User-Controllable RL Policies: Online Behavior Modification in Creative Tasks
\title{Online Behavior Modification for \\ Expressive User Control of RL-Trained Robots}

% <concept_id>10003120</concept_id>
% <concept_desc>Human-centered computing</concept_desc>
% <concept_id>10010147.10010257.10010258.10010261</concept_id>
% <concept_desc>Computing methodologies~Reinforcement learning</concept_desc>
% <concept_id>10010147.10010178</concept_id>
% <concept_desc>Computing methodologies~Artificial intelligence</concept_desc>
% <concept_significance>300</concept_significance>

% \ccsdesc[500]{Human-centered computing}
% \ccsdesc[500]{Computing methodologies~Reinforcement learning}
% \ccsdesc[300]{Computing methodologies~Artificial intelligence}

\keywords{human-robot interaction, user-centered learning, shared autonomy, reinforcement learning} 

\maketitle
\section{Introduction}

Real-world robots must complete tasks well and meet the needs of users. In many cases, a robot is optimized for only one of these. For instance, an industrial assembly line robot is programmed to perform a very specific task in a very specific way, typically in an isolated environment, and thus requires relatively little supervision from a person. Such a robot may have learned to complete the task optimally through Reinforcement Learning (RL). However, this ``one policy fits all" approach is unlikely to work when robots are working closely with humans. There are many cases where users may wish to have a robot that can autonomously perform a task while allowing for control over some dimensions of the robot's behavior. For example, a user may want a dishwashing robot to move more slowly when cleaning their favorite mug or an assistive robot to use less force when helping with dressing. While an RL-based policy may be successful at
completing the task, it may not suit the user’s in-the-moment user needs for \emph{how}
that task should be completed.

\begin{figure}[t]
\centering
\includegraphics[width=.475\textwidth]{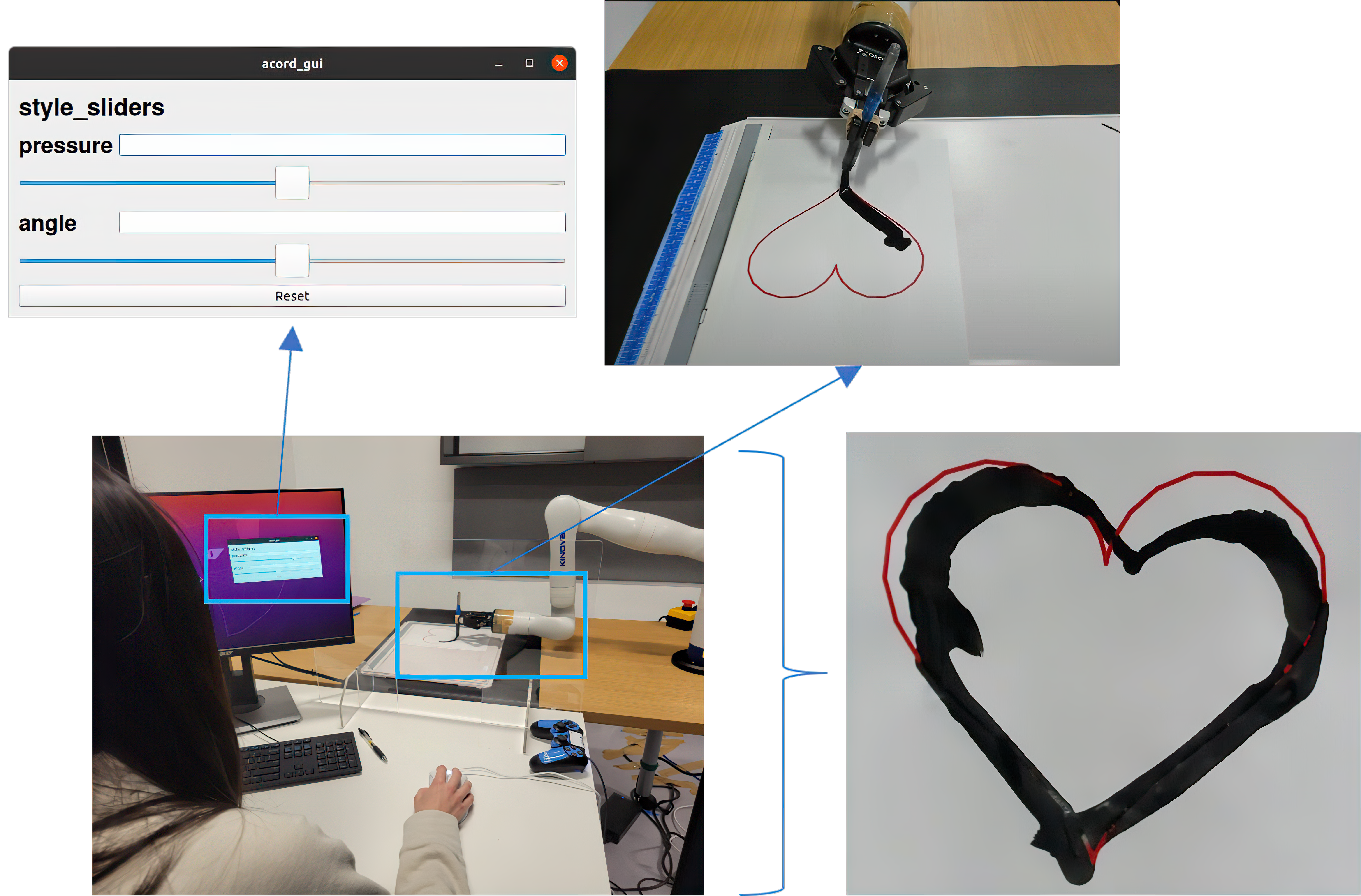}%
\captionsetup{
indention=1.5em
}
\captionof{figure}{A participant using \algoshort{} to adjust the style of a painting as the robot traces a heart autonomously.}%
\vspace*{-6mm}
\label{fig:introfig}
\end{figure}

In many situations, there is a need to facilitate interactions that give users this control over the style of task completion without burdening the user with potentially-lengthy human-in-the-loop teaching \cite{christiano_deep_2017, wu_survey_2022}. Some existing methods augment a typical RL policy, but these methods have not been adapted for or validated with real users. Existing methods include goal-conditioned RL (GCRL) for example, in which a robot's behavior is dictated by the parameterization of a goal state. Similarly, behavior diversity approaches often parameterize a robot's policy with a latent variable that encodes a skill or certain way of completing a task. These approaches are \emph{robot-centered} approaches which do not explicitly allow user control over the resulting policies. We believe these approaches can be reformulated in a \emph{user-centered} way to give users control over a robot's behavior as it completes a task. To enable close user-robot collaboration, we propose and study an approach that gives users direct control over these latent variables to adjust a robot's behavior to their liking. 

In this paper, 
we present \emph{online behavior modification}, a formulation that combines fully autonomous task completion with user-controlled behavior styles. This formulation is compatible with state-of-the-art methods for learning a task offline, such as RL, GCRL, and quality-diversity (QD), while making explicit a degree of online user control. We then present \algofull{} (\algoshort{}), a user-centered, diversity-based algorithm which serves as a proof of concept for this formulation.  We deploy \algoshort{} in a user study to demonstrate its potential in a user-centered interaction that does not sacrifice task completion. Our contributions include:
\begin{enumerate}
    \item We propose the \emph{online behavior modification} formulation, which describes an interaction where a robot autonomously completes a task while a user controls how it does so.
    \item We present \algoshort{}, a diversity-based algorithm designed for online behavior modification. \algoshort{} grants users continuous control over pre-specified behavioral features of the robot while ensuring autonomous task completion.
    \item We validate \algoshort{} in an in-person study with non-expert users (n=23) in a collaborative painting task where users adjust the \emph{style} of painting using \algoshort{}, fully-autonomous RL, and a modified version of Shared Autonomy (SA). We find that \algoshort{} is rated by users as affording the same high level of control as SA (82\% agree with ACORD giving control, 73\% with SA, versus 30\% agreeing with RL giving control), while maintaining better overall task performance (\textbf{BF}=11.67 in our measure of consistency). Furthermore, we find users strongly prefer interacting with \algoshort{} over the RL baseline (e.g., 83\% preferred ACORD, \textbf{BF}=17.16).
\end{enumerate}
\section{Related Work}
\emph{Users like to have control over robots.} Users desire this control via teleoperation \cite{passenberg_survey_2010, kebria_control_2019, kent_comparison_2017, ciocarlie_mobile_2012, walker_human_2020}, via dictating which actions the robot should \emph{not} take \cite{van_waveren_correct_2022, brawer_interactive_2023, alshiekh_safe_2017}, or via having a degree of direct control over a collaborative algorithm \cite{gopinath_human---loop_2017, zurek_situational_2021, nemlekar_transfer_2023, xing_toward_2021}. A theme of these works is that allowing users to influence robot behavior allows for more expressivity and robustness than a single policy may represent. In fact, RL is also moving towards more expressive and capable policies. Goal-conditioned RL (GCRL), for example, enables a policy to perform different tasks depending on how a goal state is selected \cite{margolis_walk_2022-1, eysenbach_contrastive_2022,liu_goal-conditioned_2022, chane-sane_goal-conditioned_2021}. Skill learning and diversity-based approaches \cite{eysenbach_diversity_2018, mandlekar_roboturk_2018,mees_adversarial_2020}, such as Quality-Diversity (QD) \cite{tjanaka_pyribs_2021, pugh_quality_2016, cully_autonomous_2019, fontaine_differentiable_2021, tjanaka_approximating_2022}, allow a robot to autonomously learn meaningfully variations in its behavior. Algorithmically similar to our work, \citet{kumar_one_2020} and \citet{osa_discovering_2022} use diversity-based approaches to increase an agent's robustness to its environment, while we propose using similar techniques to give users more control over a robot's behavior.  We also highlight that most previous robot-centered approaches have not been validated with users, a critical step to ensure that these methods serve the needs of users.

While pure teleoperation maximizes user control, our work is applicable in tasks where direct user control is impractical or impossible. A more analogous method, used as a baseline in this work, is Shared Autonomy (SA), which starts with user direct control and adds an automated assistance behavior to direct the robot to an inferred user goal or skill based on some input \cite{pitzer_towards_2011, gopinath_human---loop_2017, javdani_shared_2015, selvaggio_autonomy_2021, mower_skill-based_2021}. Although RL has also been used to enhance SA \cite{fontaine_differentiable_2021, reddy_shared_2018}, it has not been used to adjust \emph{how} the robot completes its task given the target task is known. There is a need for approaches such as ours that do just that. 

  Human-in-the-loop learning has offered approaches to guiding robot behavior via reward shaping \cite{biyik_active_2020, biyik_learning_2022, ratner_simplifying_2018, myers_learning_2022}, ensuring various safety constraints are met \cite{van_waveren_correct_2022, alshiekh_safe_2017, lutjens_safe_2019}, or, most closely related to how a robot does a task, via queries about behavior features \cite{bobu_feature_2021, basu_learning_2018, biyik_active_2020}. Approaches such as Interactive RL emphasize teaching a robot in real-time as it adapts to the teacher's feedback \cite{arzate_cruz_survey_2020, knox_interactively_2009, sheidlower_keeping_2022}. While these approaches are effective at allowing users to alter robot behavior, they often require both lengthy teaching times and retraining when a user changes their preferences, so the adjustment occurs over several executions of the task. To complement these more time-consuming methods, there is a need for approaches such as ours that allow users to quickly change a robot's behavior in the moment, within the same task execution.
% \elaine{1. humans like control}
% \elaine{2. in fact, RL is also moving towards not one policy fits all (GCRL and QD).  our approach builds off of these}
% \elaine{3. the ultimate in human control is teleoperation, next best thing is SA; we use SA as a best-case baseline for user control}
% \elaine{4. also there is HIL, but that takes forever and isn't really controllable quickly}

\section{Learning Policies for Online Behavior Modification in RL Settings}
\label{sec:algorithm}

To enable human-centered control over how a robot complete its task, we propose three key properties for \emph{online behavior modification}. First, the robot must always \textbf{autonomously make ``task progress"} and ensure the task does not fail. In this context, ``progress” may mean ``expected completion in finite time” or ``always getting closer to a goal”; formalization depends on the task. Second, there must be a non-empty set of \textbf{behavior features}, each of which has an associated \emph{behavior oversight parameter}, $k$, that control the robot along the behavior feature axis. In other words, the policy must be explicitly parameterized with one or more observable variables that dictate an aspect of the robot's behavior. Finally, for each behavior feature that has a certain $k$ associated with it, the adjustment of that $k$ must be \textbf{interpretable to a user} and there must be an \textbf{accessible interface} that facilitates a user to freely adjust each $k$ as the robot completes its task. These properties describe an interaction that ensures the user can have a robot that both meets their needs and can be personalized without having to teach the robot the task or their preferences. 

In this section, we present Adjustable Control Of RL Dynamics (ACORD), a proof-of-concept algorithm for learning a policy for online behavior modification in continuous state and action space robotics tasks. ACORD is a behavior-diversity--inspired algorithm which explicitly gives users control over a robot's behavior. We describe how to adapt a standard RL setting to facilitate ACORD and demonstrate it in a simulation environment. % and deploy ACORD on a real robot for a user study in Section IV.

\subsection{\algoshort{} for Continuous Control RL-tasks}
 We assume a task modeled as a Markov decision process (MDP) with states $S$, actions $A$, transition function $T(s,a)\rightarrow s'$, and discount factor $\gamma$. To define task failure, we assume some environmental reward function $R_\text{env}$.
To this system, we introduce \emph{\kname{}s}. Assume that $S = \mathbb{R}^n$ and define the space of \kname{}s as $K = [0, 1]^m, 1 \leq m \leq n$. Consider the coordinate representation of $s = \langle s_1, \cdots, s_i, \cdots, s_n \rangle$ and $k = \langle k_1, \cdots, k_j, \cdots, k_m \rangle$. Each coordinate of $k$, $k_j$, controls a coordinate of $s$, noted $s_i$. The set of all $s_i$ that have a $k_j$ mapping to them define a set of \emph{behavior goals} for the robot, and the corresponding $i$-axes are \emph{behavior feature} axes. Any $s_i$ with no corresponding $k_j$ is a free variable whose value is not explicitly constrained by a setting of $k$. For generality, we assume the range of behavior goals is unknown prior to learning (e.g., the maximum and minimum speeds the robot can move while completing its task are unknown). After learning, a user can directly adjust the values of $k$, thus changing the robot's behavior goal on the axis $s_i$, and consequently changing its behavior along that axis within a range that is learned by the algorithm, subject to ``non-failure'' condition above. This representation could be trivially extended to having $k_j$ control multiple coordinates. 

Learning a policy for \algoshort{} entails finding a policy parameterized by $k$, $\pi_k$, which both makes progress in the task and enforces the behavior goals. To ensure that the learned mapping from each $k_j$ to each $s_i$ is interpretable by a user, we propose the soft constraint that the robot should learn a monotonic mapping from $k_j$ to $s_i$ and that the mapping range is as large possible without preventing the robot from completing its task.

\subsection{ACORD Algorithm}
ACORD makes use of three components: a discriminator that learns a continuous mapping from $s_i \rightarrow k_j$ to generate a diversity-inspired reward; an environment reward to define failure states and a task progress heuristic $h(s,a)$ to ensure task performance; and a domain randomization component that ensures that the agent learns and is robust to various different settings of $k$ such that $k$ may be adjusted in real time. 

\textbf{ACORD Discriminator}
We train a set of discriminators $W_j$ to predict $k_j$ given $s_i$, denoted: $W_j(s_i)\in[0,1]$. We parameterize the discriminator as a neural network and train it via the novel loss function:
\vspace{-2mm}
\begin{multline}
       L(W_j(s_i), k_j) =  MSE(W_j(s_i), k_j) + \\ {\frac{1}{|\max(W_{j,s_i \sim D}(s_i))-\min(W_{j, s_i \sim D}(s_i))| + \varepsilon}}
\end{multline}
where $W_{j, s_i \sim D}$ refers to the discriminator output of a batch sampled from a replay buffer $D_W$, and $\varepsilon$ is a small number to avoid division by zero. This loss function enforces high prediction accuracy (via MSE) and that the predictions cover as wide a range as possible. The latter property is explicitly enforced by the denominator, leading to a faster convergence to the range covered by each $k_j$, resulting in more stable task behavior (see supplementary material for ablation study). 
% It does so as the robot learns and discovers new areas of the state space. This can be thought of as a type of implicit curriculum learning as explicitly done in prior work \citep{margolis_rapid_2022}. 

\textbf{RL Task Description and Agent}
We define the state space of the RL agent to be $S\bigcup K$. This makes $k$ observable to the agent. We will still denote any given state with $s$. We design a reward function such that the agent avoids failures, makes progress, and learns to enforce behavior goals:

% \setlength{\textfloatsep}{5pt}
% \begin{wrapfigure}{L}{0.8\linewidth}
% \begin{minipage}[t]{\linewidth}
% \begin{multline}
% \label{eqn:acord-reward}
%   R(s,a) =
%   \begin{cases}
%    $R_{Env}(s)$ & \text{if $s\in F^*$} \\
%    $-c$ & \text{if $h(s_{t}, a_t)$} \\ 
%    & \text{\quad  \quad $\leq 0$}
%    &\frac{1}{m}\sum_{i=1}^m(-\log|q(k_i|s_i)-k_i|) & \text{else}
%   \end{cases}
% \end{multline}
%     \end{minipage}
%   \end{wrapfigure}

\vspace{-3mm}
\begin{equation}
\label{eqn:acord-reward}
  R(s,a) =
  \begin{cases}
   R_\text{env}(s) & \text{if } s\in F^* \\
    -c & \text{if } h(s, a) \leq 0 \\
    \frac{1}{m}\sum_{i=1}^m(-\log|W_i(s_i)-k_i|) 
 & \text{else}
  \end{cases}
\end{equation}
where $R_{Env}$ denotes the reward from the environment, $F^*$ is the set of failure states which lead to a large negative reward, $h(s, a)$ denotes a heuristic for measuring task progress, and $c$ is a positive constant that punishes the agent if it fails to make task progress. 
% In a sparse reward environment, that condition can be omitted and then either a heuristic may be used or the robot's progress to the goal can be mapped directly to $k_i$ to ensure the agent can complete the task. 
Last is the reward generated by the discriminator which ensures that, for a given $k_i$, the agent is acting in the part of the state space where the discriminator can easily predict the $k_i$ value. Since $|W_i(s_i)-k_i| \in [0,1],$ this reward is always positive and the other conditions are always negative. This allows the reward function to be adapted and scaled to different environments with relative ease. Each of these terms may be scaled by a constant. We maximize this reward via the off-policy RL algorithm SAC \citep{haarnoja_soft_2018}.

\textbf{Domain Randomization Over K}
We employ domain randomization \cite{tobin_domain_2017, muratore_domain_2018} for the setting of $k$ during training. Every $n$ time steps, we sample $k_i \sim \text{Uniform}(0,1) \forall k_i\in k.$ The choice of $n$ can be difficult as when a given $k_i$ changes, it may take several steps for the robot to adjust its behavior accordingly. If $n$ is too small, the algorithm cannot learn to enforce the value of $k$ over time, and if $n$ is too large, it cannot learn to react efficiently to a user changing $k$ real time. Empirically, we find in the tasks in this paper that a reasonable choice for $n$ is about half the length of an episode; we expect that this would be the case for many tasks.

\vspace{-2mm}
\setlength{\textfloatsep}{5pt}
\setlength{\algomargin}{1em}
\begin{algorithm}\caption{ACORD}
    \label{alg:acord}
    Initialize off-policy RL Learner $\Psi$ \\
    Initialize Discriminator(s) $W$ \\
    \For{environment step t}{
    \If{nth step}{
    $k \sim \text{Uniform(0,1)}^m$
    }
    $s_t \sim s_{t,\text{env}} \text{ concatenate } k$\\
    $a_t \sim \pi_\Psi(a_t|s_t)$\\
    $s_{t+1,\text{env}} \sim p(s_{t+1}|s_t, a_t)$ \\
    $s_{t+1} = s_{t+1,\text{env}} \text{ concatenate } k$ \\
    $r_t \sim R(s,a)$ [see Eq. 2]\\
    $D_\Psi \leftarrow D_\Psi \bigcup (s_t, a_t, s_{t+1}, r_t)$\\
    $D_W \leftarrow D_W \bigcup (s_t, a_t, s_{t+1}, r_t)$\\
    \If{zth step}{
    Update $\Psi$ via gradient descent
    }
    \If{vth step}{
    Update all $W$ via loss in Eq. 1
    }
    }\end{algorithm}

\begin{figure*}[t]

\centering
\includegraphics[width=.45\textwidth]{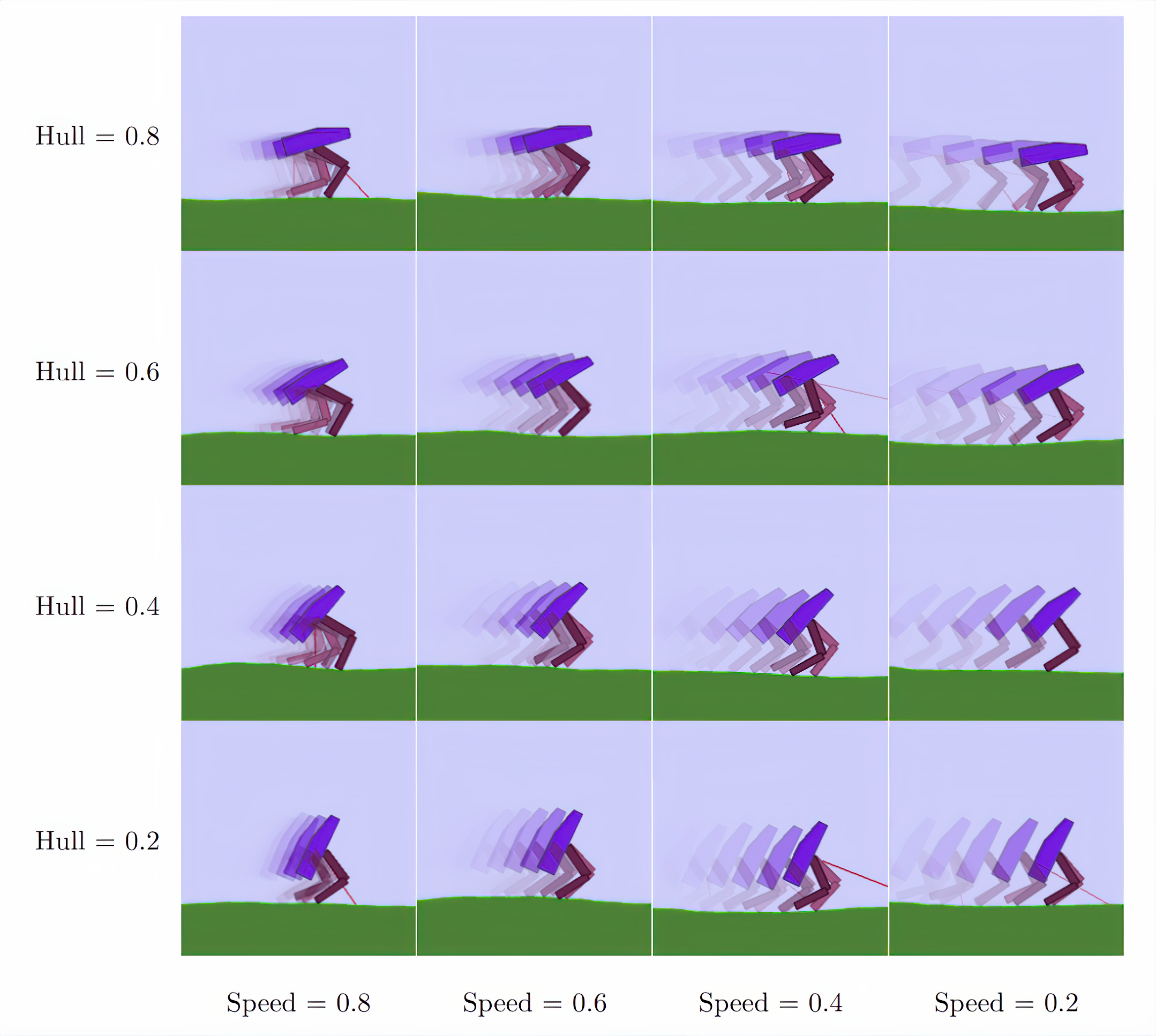} \hfill %\hspace{0.1\textwidth}
\includegraphics[width=.5\textwidth]{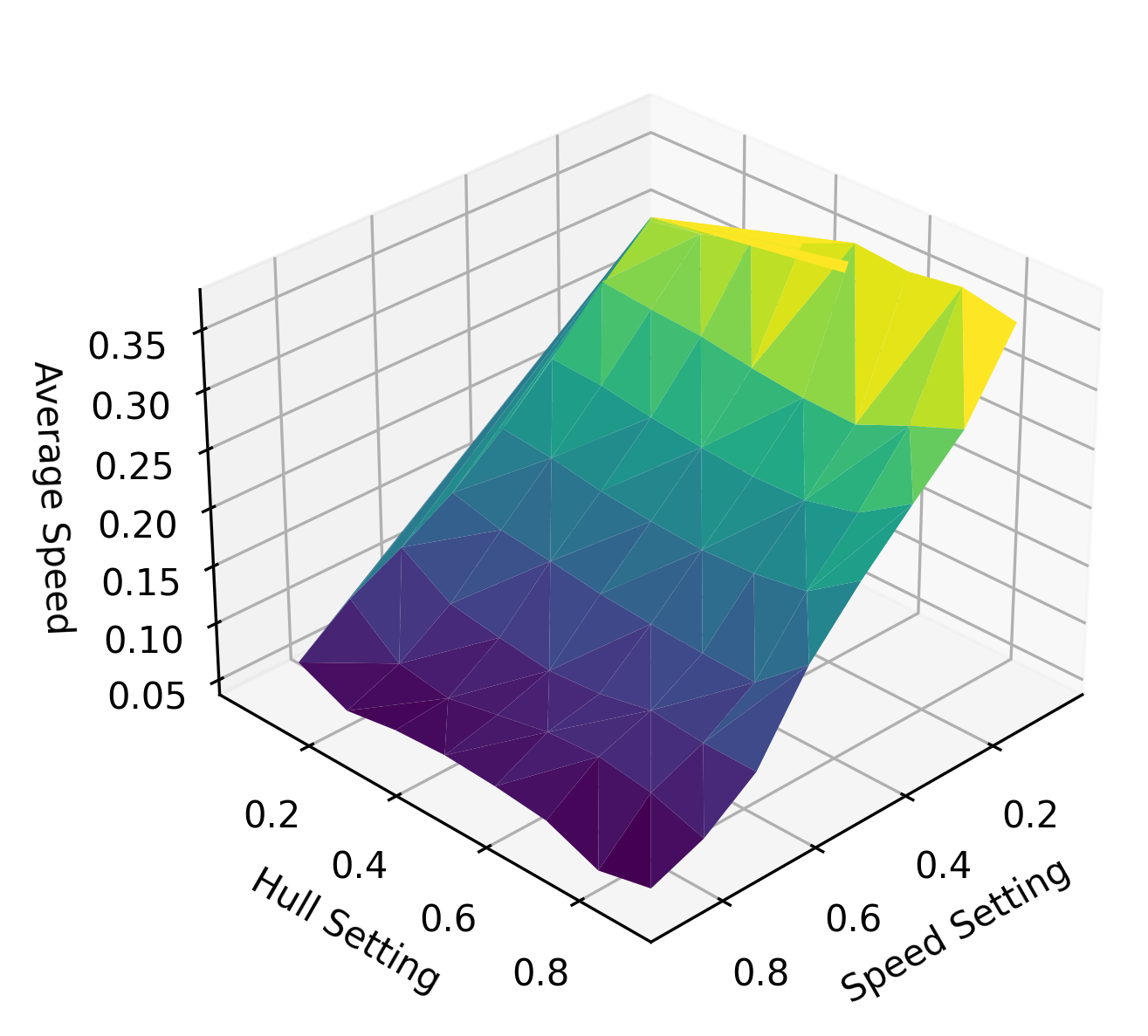}\hfill

 \vspace{-3mm}
\caption{\textbf{Left}: The walking agent varies its behavior in a predictable
and interpretable way given changes of k. The ghost traces from the previous six video frames show the agent’s change in speed. \textbf{Right}: The resulting manifold learned by ACORD in the walker environment. The speed is robust to different hull angles. }
\label{fig:sim}
 \vspace{-4mm}
\end{figure*}

\vspace{-4mm}
\subsubsection{On Using a Heuristic Progress Function}
Online behavior modification as an interaction emphasizes that the robot can autonomously complete the task by constantly making progress in that task. There are several ways to formalize this constraint, and online behavior modification does not necessarily require a particular one. For example, in this work we define a task progress measure $h(s,a)$ and require that $\pi_k$ prioritize trajectories that make $h(s,a)$ non-negative; this approach is appropriate for many robotics problems where there is a physical destination for the robot's motion (e.g., \cite{margolis_walk_2022-1}). Another natural approach might be to use the environmental reward function $R_\text{env}(s, a)$ to measure task progress or require that the trajectories following $\pi_k$ eventually reach a terminal success state. The exact specification will depend on the task and the formulation of the learning problem.

A heuristic progress function $h$ can ensure the robot always completes the task despite a user changing how it does so.  This aligns with our goal of giving users the most control possible over a robot's behavior while still accomplishing the task.  This is in contrast to prior approaches that optimally solve for a trade-off between environmental reward and diversity, as in Quality-Diversity-based approaches [24, 25], or use a hyperparameter to dictate how each of the two objectives are weighted [26]. 

\vspace{-3.5mm}
\subsection{\algoshort{} in Simulation}
We train ACORD in simulation to show that the learned policy has the desired properties: it aligns pre-specified behavior features to the values specified by $k$; it has an interpretable behavior range over $k$s; and it completes the task and avoids failures robustly in variations in $k$.
In a bipedal walker task \cite{brockman_openai_2016}, we specify two behavior oversight parameters: $k_1$ to control the speed of the robot along its $x$-axis and $k_2$ to control the angle of its hull. Failure cases are specified as crashing (-100 reward from the environment). We measure task progress by setting $h(s,a) =  v_x$, the velocity of the robot along the $x$ axis. Then, Eqn.~\ref{eqn:acord-reward} penalizes the system for moving backwards in $x$. We trained the agent to convergence prior to evaluation ($\sim$2 million steps; for a discussion of algorithm efficiency see Section 6). Figure \ref{fig:sim}, left, shows the resulting behavior by varying both $k$s. By changing $k_j$, there is a predictable change in behavior along the specified feature axis. Figure \ref{fig:sim}, right, shows the range over the robot's speed for various settings of $k_1$ given across different values of $k_2$. This demonstrates that \algoshort{} can be robust to multiple settings of $k_1$ given $k_2$: varying the hull angle does not fully constrain the agent's ability to vary its speed. Of course, if two features are directly in conflict with each other, such as a $k_i$ mapped to going backwards and a $k_j$ mapped to going forwards, the behavior of the robot may not be as expected. Lastly, over multiple runs, the agent avoids crashing $\sim94\%$ of the time with variations in many settings of $K$.

\begin{figure*}[t]
  \includegraphics[scale=.18, width=\textwidth]{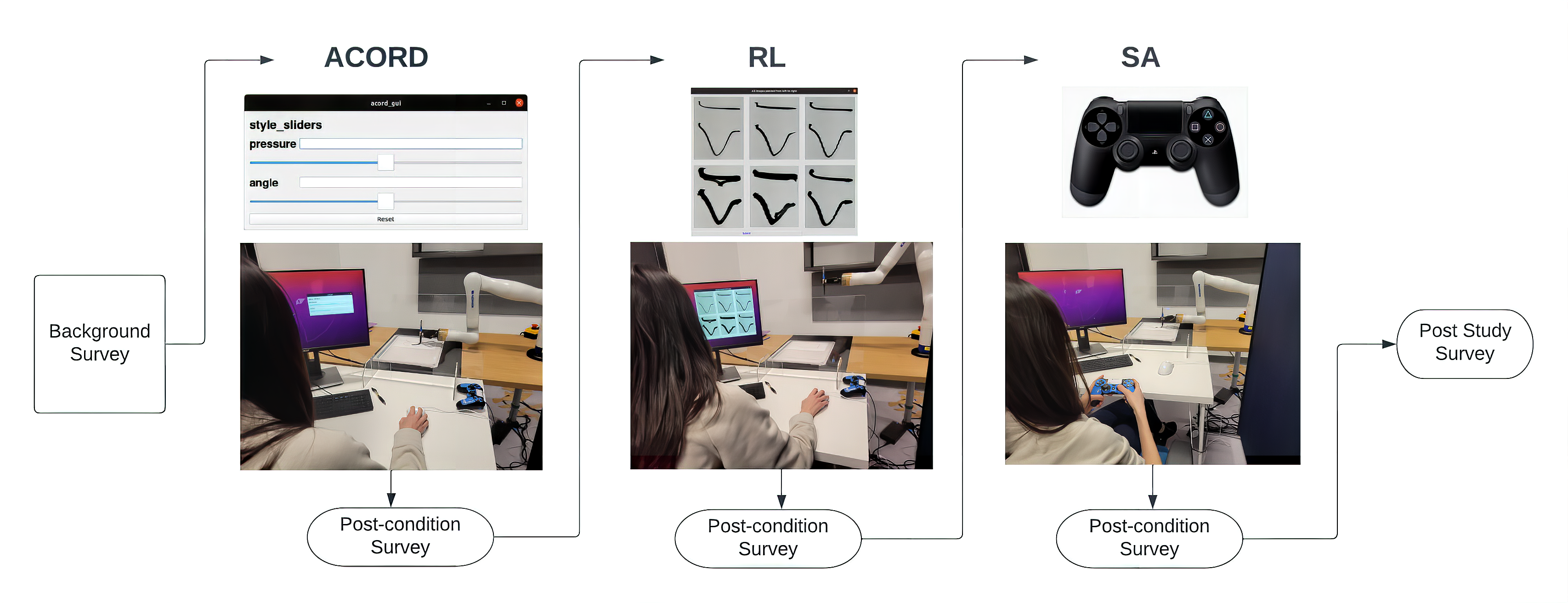}
  \vspace{-10mm}
  \caption{Overview of the study procedure.  Participants interacted with each of the three conditions (order was counterbalanced), completing a survey after each condition.}\label{fig:study}
  \vspace{-4mm}
\end{figure*}

\vspace{-4mm}
\section{User Study}
To study \algoshort{} and online behavior modification with real users, we designed a robot painting environment wherein users can adjust a robot's painting style as or before it traces a drawing. This domain is an inherently creative activity in which a person has styles and preferences that they wish to express. Online behavior modification captures the idea that task completion itself is not always the only desirable metric of a human-robot interaction: having control over \emph{how} the task is completed can also be an important factor, as is the case with painting and other artistic tasks.

\textbf{Robot Painting Task}
The painting task involved the robot tracing a previously generated shape. We specify each shape as an ordered list of waypoints in the $x$-$y$ plane, $(p_0, \cdots, p_r)$. We formulate the task as an MDP where the state $s$ is a vector containing the robot's end-effector position, orientation, and velocity; the position and orientation of a brush the robot is gripping; and the next waypoint that the robot should reach. Actions are relative Cartesian x-y velocities. Reward is given as
$R(s,a) = -|p_\text{brush}-p_i|$, 
the negative distance between the current pose of the brush $p_\text{brush}$ and the next waypoint $p_i$. Episodes terminate when the robot has reached every waypoint that makes up the shape or with failure when the arm leaves the workspace or is in collision.

\textbf{Experimental Setup}
The setup (Figure \ref{fig:study}) consisted of a Kinova Gen3 robot arm on a table with the participant sitting next to it. Depending on the condition, users had access to a different interface to interact with the robot. On the table was paper with a shape outlined in red on which the robot would paint. The participants were told which shape they would paint: heart or house (Figure 3). These shapes contain various motions and strokes and provide scope for participants to paint in their own style.

\textbf{Painting Styles}
We define two different axes for the robot to vary its painting style. One is by adjusting the height of the brush or end-effector, thus affecting the pressure that the brush applies to the canvas. This can result in thinner or wider strokes. The other way is by rotating the robot's wrist or brush. This adjusts the angle of the brush, resulting in more varied strokes.

\vspace{-3.5mm}
\subsection{Conditions}
We assume for all conditions that the robot knows how to perform the task optimally according to the MDP formulation. We fix the painting policy across each baseline to ensure the same amount of time is spent on each painting and that the style adjustment was the primary difference between conditions. We compare \algoshort{} to two alternatives to vary the style of robot behavior: RL and SA.

\textbf{Choosing Among a Discrete Set of Style-Varying RL Policies}
This condition gave the robot the most autonomy. Participants selected one of six styles based on an example image before the robot drew the shape. Each style represented a fixed value for the pitch and height of the end-effector. The robot then painted the shape autonomously according to that selected style. This type of control, in which a user chooses between a set of RL policies, is appropriate for tasks where RL control is necessary and/or available and "styles" are well defined, such as choosing a ``risky'' or ``risk-averse'' obstacle avoidance strategy. In other cases, these pre-defined policies may have been learned via human feedback, but their execution during this single task is fixed.

\textbf{Shared Autonomy (SA)}
This condition gave the participants the most direct control. Users were given assisted velocity control over the height and pitch axes of the robot end-effector through a controller. The input was augmented with a SA assistance strategy following \citep{javdani_shared_2015,javdani_shared_2018}, with $\alpha = 0.5$ to allow the user's commands to directly influence the robot position \citep{newman_harmonic_2022}. The SA assistance infers online which of the six styles defined in the previous condition the user is intending to achieve.  

While similar to the standard goal-based SA paradigm, we note two key differences. First, the system continuously moved along the $x$-$y$ plane via the optimal policy while the user controlled the style axes. Second, rather than considering goal states to be terminal, the user continued to control the style axes for the whole trajectory and could move from one goal then to another. This approach allows for the closest comparison between ACORD and SA, but this multi-goal formulation of SA is a direction for future research in itself.

\textbf{\algofull{}}
We trained and deployed an \algoshort{} agent using sim-to-real via the Gazebo simulation environment \cite{koenig_design_2004}. Failure was defined as leaving a set workspace. We defined $h(s,a) = a \cdot (p_i - p_\text{brush})$, the component of the action in the direction towards the current waypoint $p_i$. Penalizing $h(s,a) \leq 0$, as in Eqn. \ref{eqn:acord-reward}, penalizes actions that move away from  $p_i$. 

Two $k$s were learned to allow for \emph{continuous control} over the painting style: one for the height, $k_1$, and angle, $k_2$, measured at the \emph{brush tip} rather than at the robot's end-effector. This means when a user moves the slider to adjust the brush's rotation, through $k_1$, ACORD maintains contact with the paper since $k_2$ stays the same. The users had access to a GUI with two sliders to control both $k$s. Users adjusted the sliders, affecting the robot's behavior and painting style in real time. 

\begin{figure*}[t]
\centering
  \includegraphics[width=0.85\textwidth]{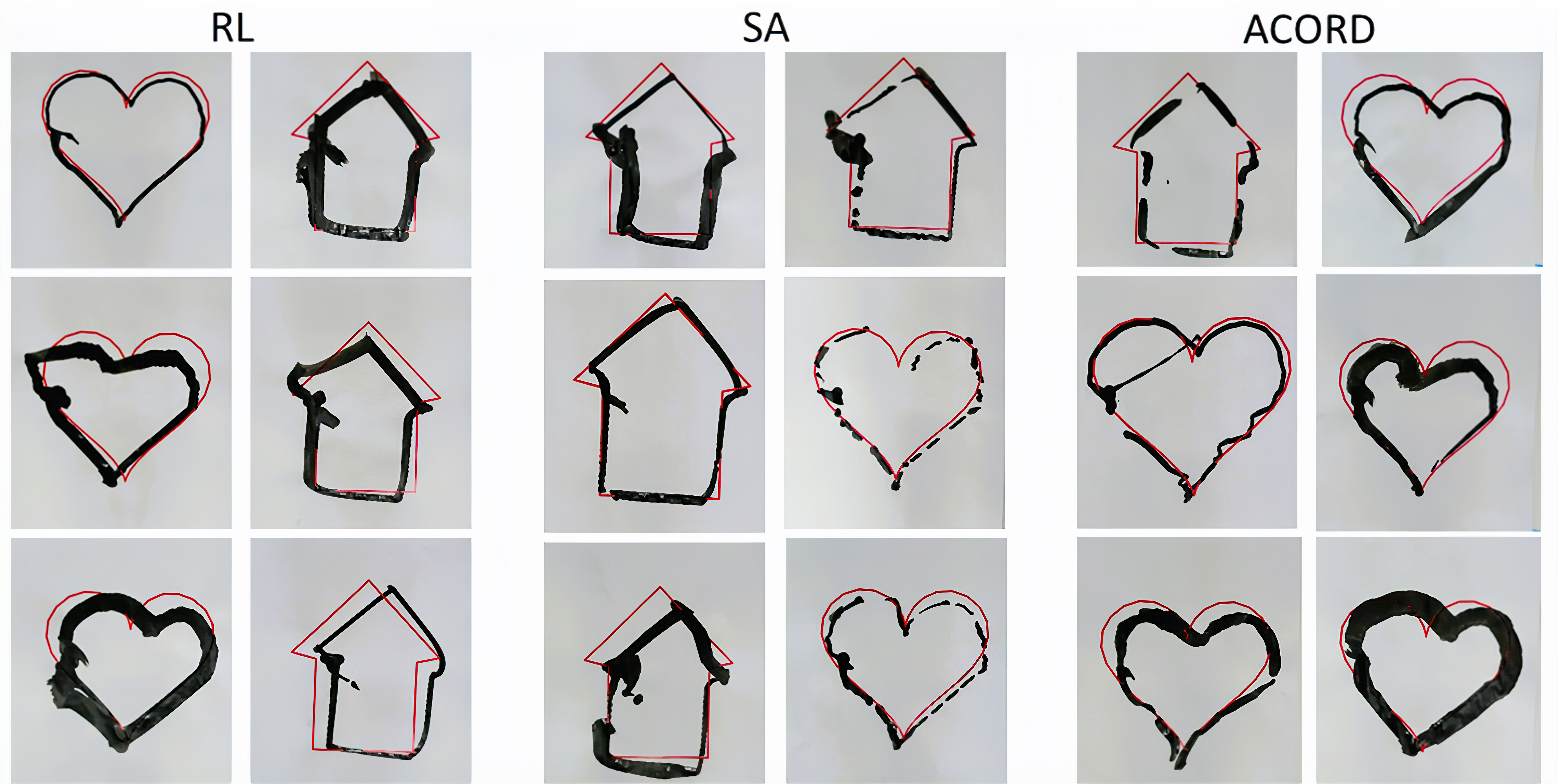}
  \vspace*{-3.4mm}
  \caption{Participant paintings. Users were able to produce a wide range of different styles for the pre-specified shapes, including the emergent ``polka dot'' style in SA (4th column from left) and widening or narrowing ``strokes'' using ACORD (rightmost column, top and center).}
  \label{fig:drawings}
  \vspace{-4mm}
\end{figure*}

\vspace{-3.5mm}
\subsection{Experimental Procedure}
\textbf{Recruitment} We recruited a total of 24 participants from the university and the surrounding area with a variety of different backgrounds. All participants were 18 years or older. Of those participants 15 were female and 9 were male. 13 participants were in the age range of 18-24, 9 in the range of 25-35, 1 in the range of 35-44 and 1 in the range of 55-64.
Participants reported their level of programming expertise from 0 (none) to 10 (expert). The mean level of programming experience was 2.9 with a standard deviation of 2.3. Furthermore, 11 participants reported having experience interacting with robots, and 3 of those 11 had significant expertise (attending robotics conferences and events regularly). The study lasted approximately 45 minutes and participants were compensated $\$15$. Of the 24 participants, the data from one participant was excluded due to non-participation (ignoring the robot's behavior and providing only uniform feedback on all surveys). This left data from $n=23$ participants for analysis. The study procedure was approved by the Tufts University IRB. 

\textbf{Procedure} Participants provided informed consent then took a background survey.  The experimenter then explained the task and control in the conditions, including allowing participants to practice with SA and \algoshort{}. In each condition, participants painted the house shape and then the heart shape, then filled out a survey about that condition. Conditions were fully counterbalanced within subjects. Finally, participants completed a post-study survey, were thanked, and given compensation. 

\textbf{Outcome Measures}
The post-condition survey included NASA TLX~\citep{noauthor_tlx_nodate} and UTAUT~\citep{venkatesh_user_2003} surveys. We adjusted the scale of all questions to a 5-item Likert-scale. We also asked two other Likert-scale questions: \emph{I had control over the robot’s behavior} and \emph{I could express myself through the robot}, and an open response question: \emph{How much do you feel the robot's ability to complete the task depended on your input?} The post-study survey had participants rank each condition based on their preference, the ability to express themselves, the perceived reliability of how well the robot traced the shape, and which mechanism (e.g. controller or sliders) they preferred. In addition, it asked two open response questions: a request for general comments and the question \emph{how could the interactions be improved?}% and provided open responses to a request for general comments  and how the interactions could be improved.

We evaluated two quantitative metrics for how reliably the shape was traced. For each painting, we calculated the \emph{coverage}, or percentage of the red line that remained visible in the image after the task was complete. We also calculated the \emph{consistency}, or the coverage of the red line after applying translations and rotations of the painting to best align with the shape of the red line.

\textbf{Hypotheses} We expect that ACORD will give users control over the robot's behavior while still effectively completing the task, as users have more direct control than RL but less than that of SA.  Thus, we expect that ACORD will be the most preferred approach and that it will give users feelings of slightly less control as SA while having similar performance to RL. This results in three hypotheses:

    \textbf{H1:} Users will prefer to interact with \algoshort{} over SA and RL.
    
    \textbf{H2:} Users of \algoshort{} will feel more in control of the robot than in RL but less than in SA.
    
    \textbf{H3:} RL will be objectively and subjectively the most reliable, \algoshort{} the second most and SA the least. 
\vspace{-2mm}

\begin{figure*}[t]
    \centering
    \includegraphics[width=\textwidth]{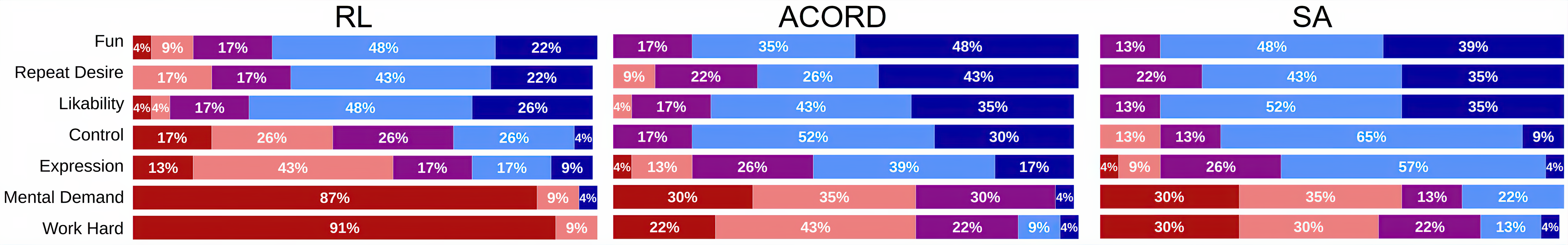}
    \vspace{-7mm}
    \caption{Responses to post-condition 5-point Likert scale questions. The darkest blue represents "strongly agree" or, in the case of Mental Demand, "very high." The darkest red represents "strongly disagree" or, in the case of Mental Demand "very low."}
    \label{fig:likert}
    \vspace{-3mm}
\end{figure*}
\setlength{\textfloatsep}{5pt}

\begin{figure*}
    \centering
    \includegraphics[width=.79\textwidth]{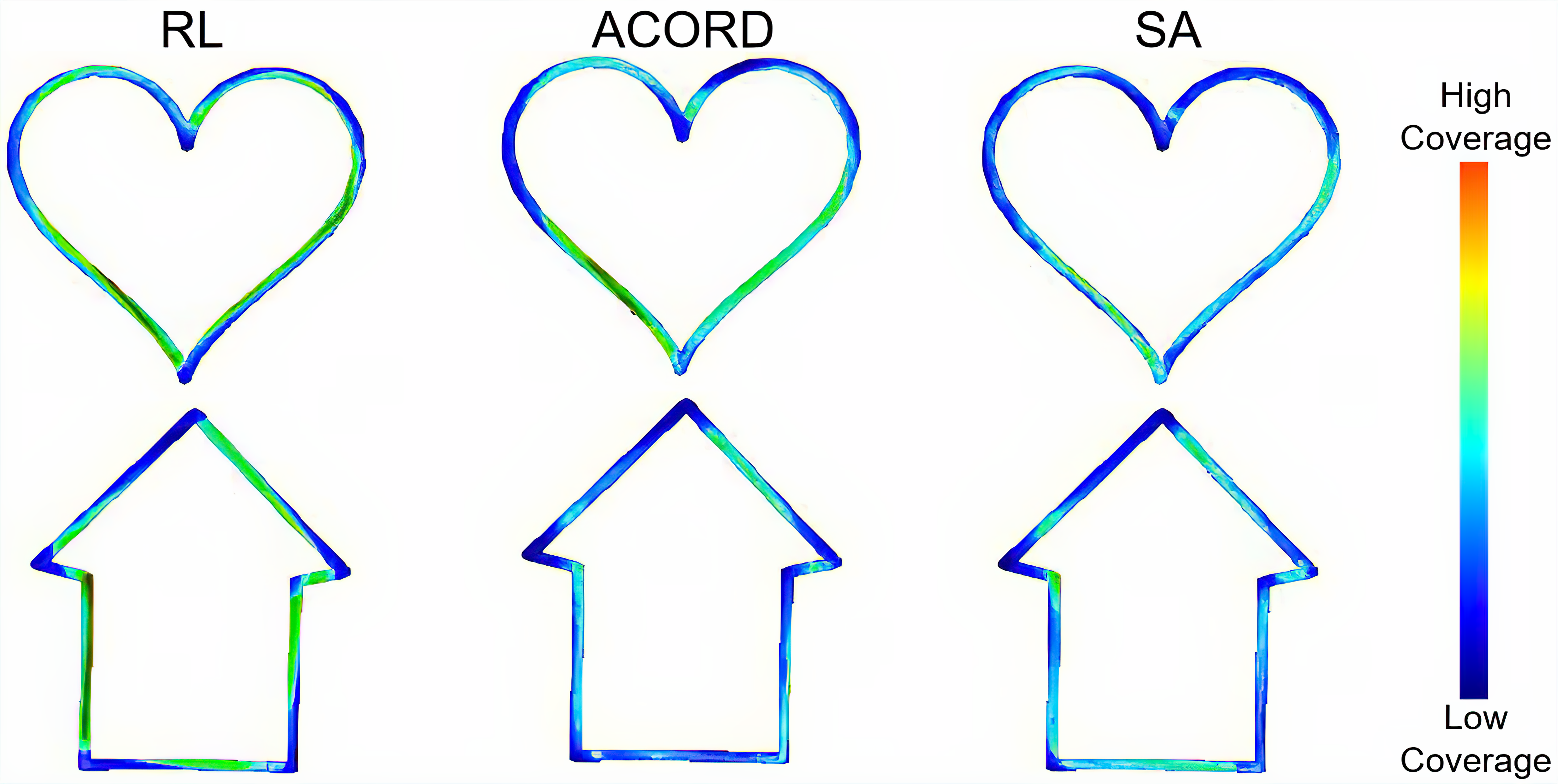}
    \vspace{-3mm}
    \caption{Heatmaps depicting the \emph{consistency} of each approach sorted left to right from most consistent overall to least consistent. The heatmap consists of the participant's paintings layered on top each other after being shifted for maximal coverage. Areas of high coverage depict areas where many participants painted over, and vice versa for areas of low coverage.}
    \label{fig:heatmap}
    \vspace{-4mm}
\end{figure*}
\setlength{\textfloatsep}{5pt}

\section{Results}
%\textbf{Data coding and Analysis}
To analyze the data, we use Bayesian statistics following the interpretation scheme presented in \cite{van_doorn_jasp_2021}: a Bayes Factor (BF) between 3 and 10 we interpret as ``moderate evidence'' for the alternative hypothesis, between 10 and 30 as ``strong evidence,'' and 30 or above as ``very strong evidence.'' To evaluate the post-study survey data, we encoded responses as pairwise comparisons between two of the three conditions. For each comparison, the rank was encoded as 1 if the ``left'' condition was preferred, -1 if the ``right'' condition was preferred, and 0 if the participant ranked the two conditions equally. To analyze this data, we used a Bayesian Wilcoxon Signed Ranked test with a Cauchy prior distribution with $r=1/\sqrt{2}$. To analyze the Likert scale data , we used a Bayesian Repeated Measures ANOVA. We used a Bayesian Paired Samples T-Test to analyze the coverage and consistency metrics.

% \begin{figure*}[t]       
%     \fbox{\includegraphics[scale=.33]{plots/Likert/like_repeat_plot_v2.png}}   
%     \hspace{0px}
%     \fbox{\includegraphics[scale=.33]{plots/Likert/contExpress_plot_v2_nolegend.png} \includegraphics[scale=.33]{plots/Likert/mental_hard_plot_v2.png}}
%     \caption{this is the caption}
%     \label{materialflowChart}
% \end{figure*}

\textbf{User preferences}
We find strong evidence that \algoshort{} is preferred over RL (\textbf{BF}=17.16) and anecdotal evidence that people prefer SA over RL (\textbf{BF}=2.11). There is strong evidence that people found \algoshort{} more fun than RL (\textbf{BF}=79.87) and moderate evidence people found SA more fun than RL (\textbf{BF}=5.03). These results provide support for \algoshort{} being preferred over RL  while being no less preferred than SA. We also find a trend towards \algoshort{} being preferred to a greater extent over RL than SA.  Finally, we found that users rated RL as much less mentally demanding than SA and \algoshort{} (\textbf{BF}=112.87 and \textbf{BF}=45.92 respectively), and much less hard work (\textbf{BF}>10000 and \textbf{BF}>10000), although the previous results suggest this was not a significant factor in user preferences. These findings partially support \textbf{H1} and directly support that \algoshort{} provides at least as much benefit to user experience as SA.

% \begin{figure}[t]

% \centering
% \includegraphics[width=.9\textwidth]{plots/likert/likert_graph_corl_v3.png}
% \caption{Results from post-condition Likert scale questions. The darkest blue on the right indicates "strongly agree/very high," the darkest red on the left indicates "strongly disagree/very low."}
% \label{fig:figure3}

% \end{figure}

\textbf{User Control and Expression}
 In the post study-survey we find strong evidence that people find \algoshort{} and SA more expressive than RL (\textbf{BF}=18.40 and \textbf{BF}=13.65) and similarly for the post-condition survey measure of expressiveness
(\textbf{BF}=23.38 and \textbf{BF}=40.31). Users also found a greater sense of control with \algoshort{} and SA
 (\textbf{BF}=6318.61 and \textbf{BF}=40.31). There is anecdotal evidence that users reported more control in \algoshort{} than SA (\textbf{BF}=2) and differences between the two were often commented on in open-ended responses. These results support the first part of \textbf{H2},  that users felt more in control in \algoshort{} than in RL, however our results suggest that some users may have felt an even \emph{greater} sense of control in \algoshort{} than in SA.
 
 \textbf{Quantitative Painting Analysis}
 We find on average, across both shapes, \algoshort{} and SA had better coverage than RL (\textbf{BF}>10000 and \textbf{BF}=1095.2), likely due to the persistent offset in the RL condition caused by \emph{bristle drag} of the brush. We account for misalignment by computing the maximum coverage found over small translations and rotations of the template, which we refer to as consistency. As expected, RL has better consistency than SA and ACORD in both shapes and, in general, the normalized sum across both shapes (\textbf{BF}>10000). While SA has higher consistency in the house shape (\textbf{BF}=1884.64), \algoshort{} has much higher consistency in the heart shape and a higher consistency overall (\textbf{BF}>10000 and \textbf{BF}=11.67). A visualization of the consistency results can be found in Figure \ref{fig:heatmap}. According to our two reliability metrics, \textbf{ H3} is supported by the consistency metric and not by the coverage metric. The coverage findings, however, showcased how a human in the loop can use the flexibility of added control to compensate for execution-time limitations in pre-trained RL models. 

\textbf{Qualitative Results}
Figure \ref{fig:drawings} shows paintings from each condition that are representative of the different painting styles found and the \emph{emergent behaviors} that users demonstrated. With \algoshort{}, we see the emergent behavior of brush strokes, where users moved both sliders quickly to make a specific stroke. In SA, some users made polka dots by bringing the brush up as much as they could, releasing the joystick, then letting the assistance bring the brush back to the paper. This was a surprising use of SA and goes against the task description of tracing the shape, yet gave users who figured this out a new way of expressing themselves and highlights that users had a desire for control and creativity in the task. 
While both \algoshort{} and SA enabled this control, many users emphasized ``consistency'' and ``ease of use'' when describing \algoshort{}; in contrast, users described SA as ``mentally demanding'' or ``too sensitive.''
Some users did not enjoy that \algoshort{} required "shifting their eyes" from the screen to the robot, although of course this is an issue with the interface and not with \algoshort{} itself. RL was criticized for not being able to adjust the style in real time; however, multiple users said it would be ideal for a "mass production" setting.

\begin{figure}[t]
    \begin{subfigure}{.4\textwidth}
    \centering
    \includegraphics[scale=0.5]{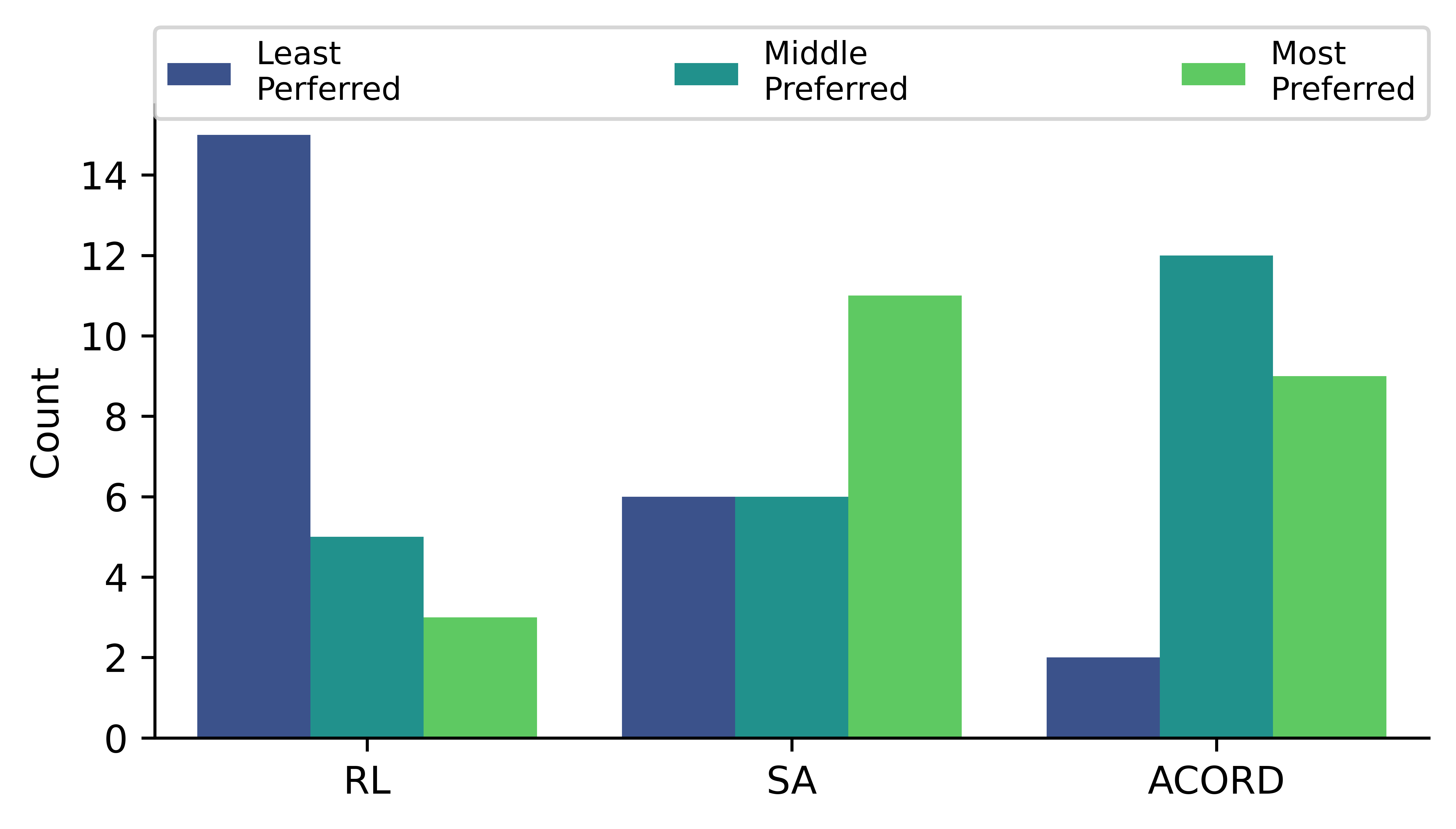}
    \label{subfig:a}
    \end{subfigure}
    \begin{subfigure}{.4\textwidth}
    \centering
    \includegraphics[scale=.5]{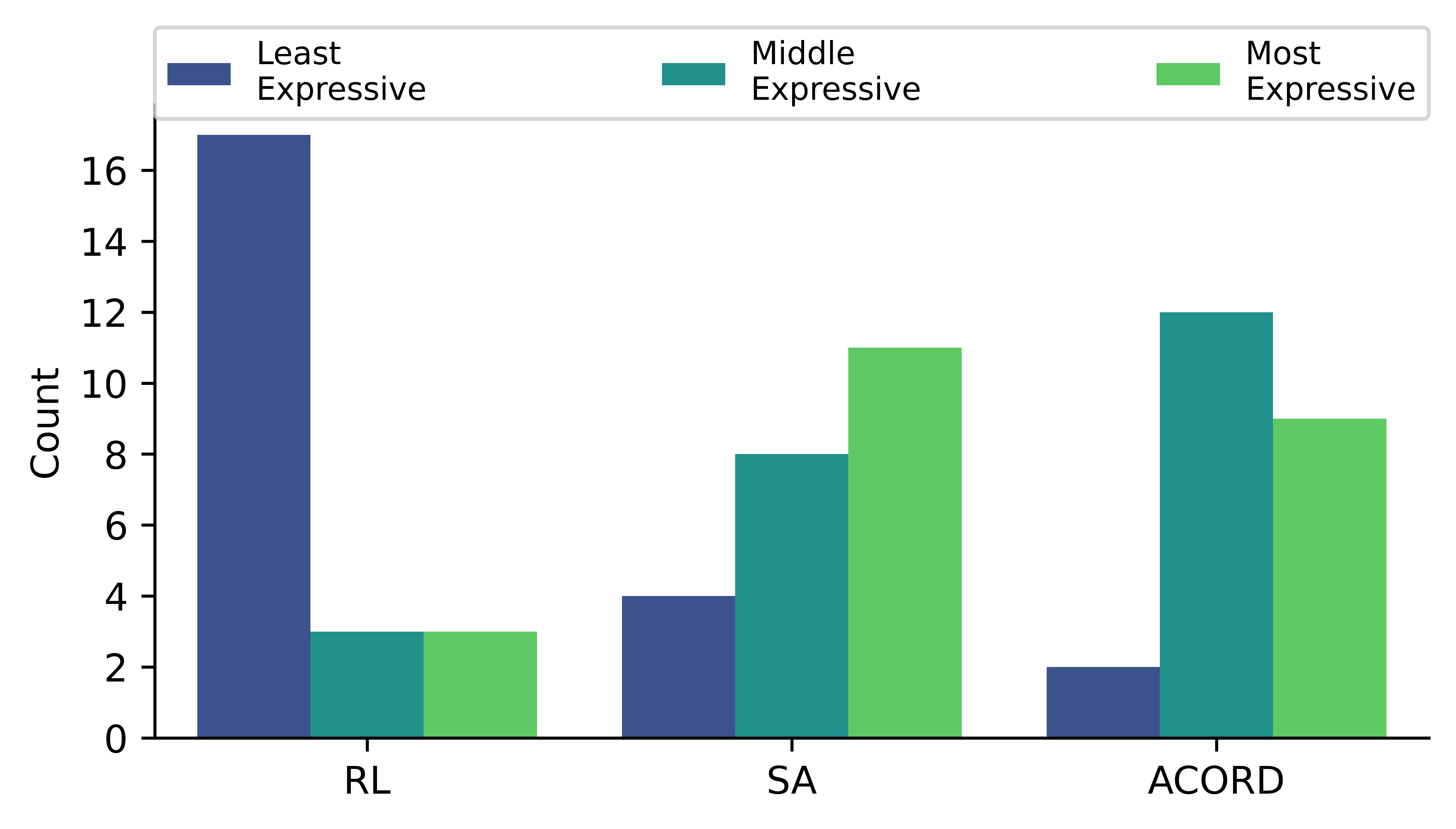}
    \label{subfig:b}
    \end{subfigure}
    \begin{subfigure}{.4\textwidth}
    \centering
    \includegraphics[scale=.5]{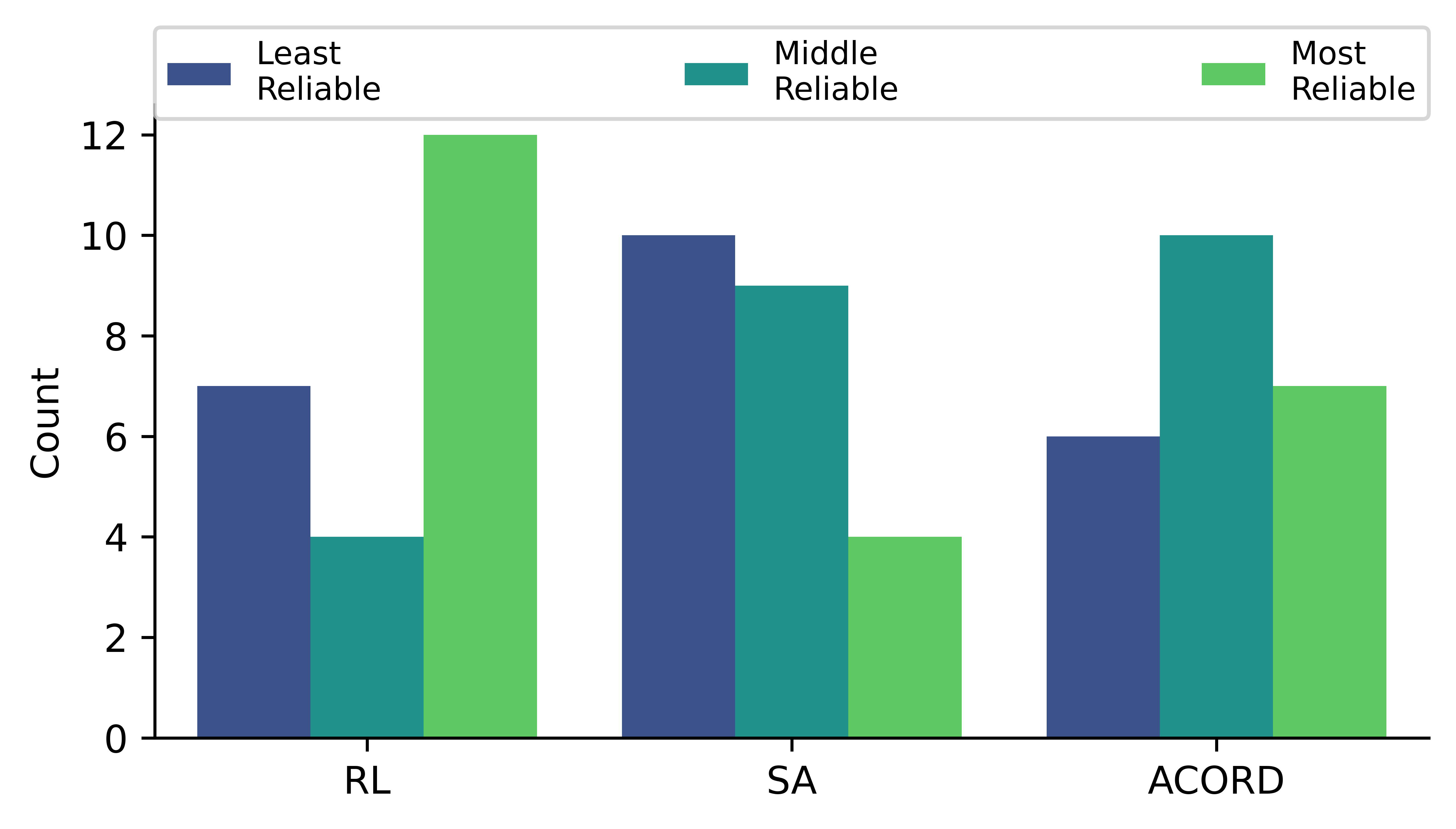}
    \label{subfig:c}
    \end{subfigure}
    \vspace*{-5mm}
    \caption{Results of the post study surveys. Users ranked each condition based one their preference (top), perceived expressive potential (mid), and perceived reliability (bottom).}\label{fig:post}
     \vspace{-1mm}
\end{figure}

\vspace{-3mm}
\section{Discussion}
Online behavior modification describes an interaction in which a user has control over how an otherwise autonomous robot completes a task. While prior work has offered various algorithmic avenues to fulfill this type of user control, such as GCRL or Skill Learning, they have been formulated in robot-centered ways and lack validation in terms of usability and acceptance by actual users. In contrast, online behavior modification is a user-centric formulation that can leverage the benefits of these approaches to empower users in ways that can be systematically tested and compared.

Online behavior modification occupies a novel place within approaches to combine autonomous execution with human input. Our user study compared the \algoshort{} algorithm to both a library of autonomous RL policies and a version of SA modified for a multi-goal setting where different styles represent different goals. We validate that ACORD can be used to adjust the style of a robots behavior and is perceived favorably by users. Our study shows that ACORD provides high levels of perceived control and expressiveness, as SA does, while being easier to use. There are also key technical and theoretical differences between online behavior modification and SA. In the context of SA, the task-level goal is unknown, and the robot, through an interpretation of the user’s control signal, is attempting to infer the goal of the task. In contrast, in online behavior modification, the task-level goal is known, and the purpose is to maximize the user’s control over how the robot autonomously completes that task. SA also requires the user to operate directly in the robot’s action-space defined for the task, while algorithms such as ACORD build a separate new space for user input. In a larger system, online behavior modification algorithms like ACORD could work \emph{with} SA, for example by using an SA system to infer \emph{where} the user wants to go, and ACORD to give the user control over \emph{how} the robot gets there. This opens up various directions for future research, both studying and comparing different algorithms for online behavior modification, as well as how online behavior modification may fit into or be combined with other paradigms.

\textbf{Limitations} An assumption in this work is that the designers of the system \emph{know which axes of behavior people care about for the task}. This could be resolved by working with users to understand which behavior features they wish to adjust. Future work might also develop a general understanding of the types of features that users most want to adjust for a given task or types of tasks. Another limitation of the study is that we only considered m=2 behavior parameters to adjust. \citet{osa_discovering_2022} have shown that the diversity-based methods ACORD is partially based on can learn effectively with up to 25 discrete latent variables. However, a large number of latent variables may impede the usability and interpretability of the system. Thus, more work is needed to understand how users interact with more numerous and abstract features. While ACORD was sufficiently efficient to be deployed on a real robot and be used by real users, the algorithm is relatively sample-inefficient (about 3 hours of fine-tuning after training in simulation). Future work could improve ACORD's efficiency by leveraging other techniques, such as hindsight and Constrained MDPs \cite{andrychowicz_hindsight_2018, altman_constrained_2021}. Lastly, although online behavior modification entails the robot avoid task failures, that specification may not be sufficient for saftey-critical scenarios unless, potentially, combined with safe RL methods \cite{alshiekh_safe_2017, garcia_comprehensive_nodate, marta_human-feedback_2022}.

\textbf{Conclusion} 
This paper introduced the online behavior modification formulation, in which a user has control over \emph{how} an otherwise-autonomous robot completes a task. Leveraging robot-centered algorithmic approaches for varying robot behavior, we proposed ACORD, a user-centered behavior diversity inspired algorithm that explicitly allows users continuous control over behavior features of a robot. We demonstrate ACORD's applicability to online behavior modification in simulation prior to deploying it in a user study.
Interacting using ACORD was strongly preferred over selecting among RL policies, likely due to its creative potential and real-time control element, while its task accuracy and ease of use outperformed SA, in addition to being usable in tasks for which SA is not appropriate. This work highlights how human-centered formulations of robot learning can be used to enhance user experience with robots and opens directions for future research in this area.
%Users not only prefer interacting with this approach in a creative task over choosing amongst RL policies, but even allowed for similar levels of control of SA without sacrificing any task robustness.

\vspace{-3mm}
\begin{acks}
    The work described here was supported in part by the US National Science Foundation (IIS-2132887). 
\end{acks}

\bibliographystyle{ACM-Reference-Format}
\bibliography{references}
\end{document}